\documentclass[runningheads]{llncs}
\usepackage[T1]{fontenc}
\usepackage{graphicx}
\usepackage{xcolor}
\usepackage{soul}
\usepackage{multirow}
\begin{document}
\title{Material synthesis through simulations guided by machine learning: a position paper}

\newcommand{\eg}[0]{e.g.}
\newcommand{\ie}[0]{i.e.}
\newcommand{\etc}[0]{\textit{etc.}}

\newcommand{\todo}[1]{\textcolor{red}{\textbf{#1}}}

\titlerunning{Material synthesis through ML-guided simulations}
%
\author{
Usman Syed\inst{1} \and Federico Cunico\inst{1} \and Uzair Khan\inst{1} \and Eros Radicchi\inst{2} \and \\Francesco Setti\inst{1} \and Adolfo Speghini\inst{2} \and Paolo Marone\inst{3} \and \\Filiberto Semenzin\inst{4} \and Marco Cristani\inst{1}
}
\authorrunning{U. Syed et al.}
%
\institute{University of Verona, Italy, Department of Engineering DIMI
\\\email{name.surname@univr.it} \and
NRG-Department of Biotechnology, University of Verona and INSTM, RU of Verona, Strada Le Grazie 15, I-37134 Verona, Italy \\\email{eros.radicchi@univr.it} ;  \email{adolfo.speghini@univr.it} \and
Istituto Internazionale del Marmo, Italy\\\email{paolo.marone@stonetechtraining.it}
\and
Verona Stone District, Italy 
}
\maketitle              
%



\begin{abstract}
    In this position paper, we propose an approach for sustainable data collection in the field of optimal mix design for marble sludge reuse. Marble sludge, a calcium-rich residual from stone-cutting processes, can be repurposed by mixing it with various ingredients. However, determining the optimal mix design is challenging due to the variability in sludge composition and the costly, time-consuming nature of experimental data collection. 
    Also, we investigate the possibility of using machine learning models using meta-learning as an optimization tool to estimate the correct quantity of stone-cutting sludge to be used in aggregates to obtain a mix design with specific mechanical properties that can be used successfully in the building industry. 
    Our approach offers two key advantages: (i) through simulations, a large dataset can be generated, saving time and money during the data collection phase, and (ii) Utilizing machine learning models, with performance enhancement through hyper-parameter optimization via meta-learning, to estimate optimal mix designs reducing the need for extensive manual experimentation, lowering costs, minimizing environmental impact, and accelerating the processing of quarry sludge. Our idea promises to streamline the marble sludge reuse process by leveraging collective data and advanced machine learning, promoting sustainability and efficiency in the stone-cutting sector.

    \keywords{marble recycling \and stone-cutting sludge \and position paper}
\end{abstract}

\section{Introduction}\label{sec:introduction}

For centuries, natural stone has been the cornerstone of building and design. Quarries and cutting techniques hold deep historical and cultural significance in many regions. This enduring legacy continues today, as natural stone offers timeless beauty and unmatched durability compared to many synthetic materials. This makes it a sought-after choice for high-end buildings, monuments, and architectural projects. However, the extraction and processing of stone, including cutting and smoothing, comes with a significant environmental cost: the production of cutting sludge~\cite{nasserdine2009environmental,zichella2021ornamental}. 
Stone-cutting sludge poses a significant ecological and economic challenge, particularly in Europe and, more broadly, in several Western Countries. This industrial byproduct is generated in substantial quantities—approximately five million tons annually in Europe~\cite{zichella2018preliminary}—and predominantly end up in landfills, necessitating costly and environmentally detrimental disposal methods. In some areas of northern Italy, particularly famous for its marble, the annual production of sludge by around 750 companies amounts to over 230 thousand tons, incurring disposal costs esteemed between 5 and 8 million euros annually. Thus, the economic advantage that comes from the bare avoidance of producing such a waste is evident. 

In this sense, the opportunity to recycle stone-cutting sludge rather than treating it as waste is compelling~\cite{tunc2019recycling,al2021recycling}. One way to virtously reuse marble sludge, with the double aim of avoiding wastes and enhancing specific material properties, is to incorporate them into concrete mixes as a partial replacement for raw materials such as cement or fine aggregate~\cite{branco2023partial}.
Marble dust is increasingly used in the construction industry as a partial replacement for sand in concrete, enhancing sustainability and reducing reliance on natural sand. When used in concrete mixtures, marble dust can improve the concrete's compressive strength, workability, and durability. It fills the voids between coarse aggregates, creating a denser and stronger mix. The problem lies in which substances should be included in the mix and in which amount. Also, the type of process, such as crushing, grinding, and sieving, which brings from the input ingredients to the output material, poses many challenges.

This problem is generally called mix-design~\cite{alzboon2009effect}, and it is the crucial material engineering phase to synthesize products that satisfy the required chemical and physical properties. The specific properties depend on many factors, such as the type of construction (roads, buildings, etc.), as well as laws, regulations, and standards, which are often very different from country to country. 
Usually, the mix-design procedure is empirical, with random and lengthy trials required during the design procedure, which can consume immense manpower and materials resources. Data collection associated with this process is, therefore, long and expensive. These costs create a bottleneck, impeding scalability and ultimately hindering large-scale marble dust upcycling. Moreover, this empirical approach complicates compliance with existing laws, regulations, and standards, as it may require numerous experiments. This need for extensive testing potentially results in an impractical situation, further obstructing efficient upcycling efforts. 

\begin{figure}[h!]
    \centering
    \includegraphics[width=0.5\linewidth]{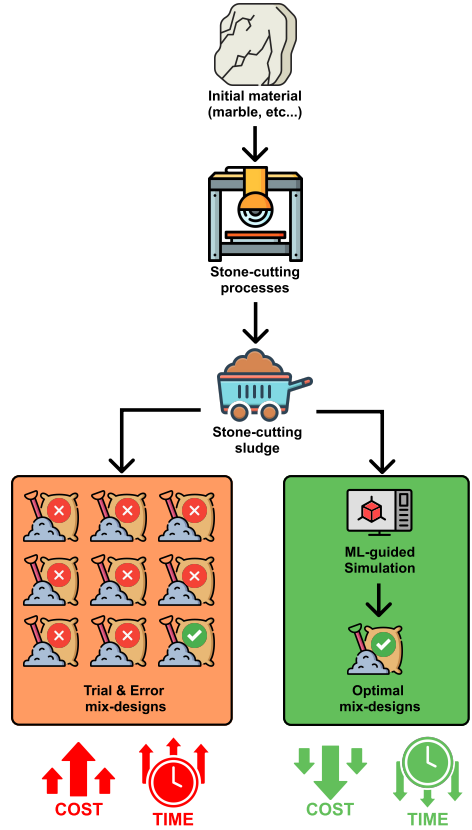}
    \caption{The stone-cutting sludge problem overview. The stone-cutting processes of marble and similar materials produce some byproducts, such as stone-cutting sludge. These byproducts come with a high cost for the environment and industries. They can be disposed of as waste, leading only to a cost for the industry, or be re-used to create mix-designs. 
    Traditionally, the reuse is subject to trial \& error approaches, involving a high cost and requiring a lot of time.
    Instead, we argue that it is possible to use machine learning-driven simulations to predict optimal mix-designs. In this way, it is possible to obtain products from byproducts that are considered waste, thus saving time, increasing sustainability, and reducing the costs for the industry.}
    \label{fig:enter-label}
\end{figure}


This position paper states that machine learning simulation can be used to collect a large amount of data that can be used to discover an optimal mix-design, particularly using meta-learning optimization techniques to identify the best parameters for the model. 
By learning from various tasks and quickly adapting to new ones, meta-learning algorithms can efficiently explore vast design spaces and identify promising configurations. In our research, we are applying meta-learning to a range of methods, including support vector machines (SVM), Random Forest (RF), Gradient boosting (GB), and XGBoost (XGB).




To prove our claim, and given the scarcity of available data on mix-designs, either due to limited research or by industrial secrets that cover the recipes of proprietary mix-designs - we applied the proposed methodology to a related problem, i.e. finding the optimal water-binder ratio in the production of concrete.  In fact, this is one of the important factors to take into account when creating concrete-based mix-designs and can serve as a good proxy for the task of creating mix-designs from stone-cutting sludge.


Specifically, we consider a dataset on concrete~\cite{cao2023prediction} consisting of 240 data records featuring 74 unique concrete mixture designs. Here, the independent variables are water-binder ratio (`WB'), binder content in \(kg/m^3\) (`Binder'), fly ash content in \% (`FlyAsh'), slag content in \% (`Slag'), superplasticizer content in \% (`Superplasticizer'), ratio of coarse aggregates to fine aggregates by weight ('Aggregate'), number of curing days (`CuringDays'), and the compressive strength (`Strength'), while the dependent variable is the porosity, which remains a significant factor in determining the quality and performance of the concrete. 
With the GB model through meta-learning, specifically Bayesian optimization, we show how we can predict the porosity with a Mean Squared Error (MSE) of 0.1077. Such error values unlock the possibility of trusting the simulation, discovering the proper mix-design, and performing, only then, a restricted number of real mix-design experiments. 

This position paper advocates for unlocking the potential of stone-cutting sludge in construction materials. We propose to achieve this through large-scale simulation campaigns, offering significant advantages: minimal cost, rapid turnaround time, and reduced environmental footprint. By adopting simulation-based techniques, we aim to improve traditional trial-and-error testing methods, thereby accelerating progress toward these goals. This strategy leverages powerful machine learning techniques, such as meta-learning, to optimize mix-designs for specific use cases. The potential of this initiative extends beyond environmental benefits; it also promises substantial economic advantages. By reducing the volume of stone sludge destined for landfills, the approach contributes to establishing a circular economy, where waste materials are repurposed and reintegrated into the production cycle. This minimizes waste disposal costs and creates a new market for recycled materials, fostering sustainable industrial practices.



\section{Related Works}\label{sec:related}

In recent years, the utilization of stone or marble dust, a byproduct of stone crushing and processing, has surged in the construction industry \cite{Silva2023}. Its growing popularity stems from unique physical and chemical properties that enhance its appeal as a versatile construction material. Characterized by excellent compaction properties and a fine granular texture, stone dust significantly bolsters the strength and stability of construction materials \cite{Song2011}. Moreover, its mineral composition, enriched with nutrients such as calcium, magnesium, and trace elements, contributes to enhanced durability and resistance to environmental degradation in composite materials \cite{Bahoria2013}.  By integrating stone dust into construction practices—either in concrete production, road base material, or as an asphalt filler—the industry reduces waste and conserves precious natural resources, fostering more sustainable construction methodologies \cite{Turuallo}. 

Ural et al. \cite{Ural2020} conducted a pivotal study evaluating marble waste as a foundational material for road construction. Its suitability was assessed by carrying out several physicochemical and mechanical tests, including freeze-thaw, Los Angeles abrasion, flatness index, water absorption,  modified Proctor test, and California Bearing Ratio (CBR), concluding that marble waste presents a robust, environmentally sustainable solution for road foundation construction, effectively managing waste within the marble industry. 

In a separate study, Jain et al. \cite{Jain2020} explored the use of marble dust to enhance the properties of subgrade soils for highway construction, finding that the inclusion of marble dust up to a certain percentage enhances the strength of black cotton soil, making it optimal to be used as a base in road construction.  Similarly, Segui et al. \cite{Segui2023} detailed the utilization of waste rock as a primary aggregate material in road construction. Whether used directly or enhanced with hydraulic binders like cement or lime, untreated waste rock bolsters mechanical strength and promotes efficient drainage in road subbase and base layers. Stabilization techniques such as geopolymerization or bituminous treatment further enhance durability, effectively reducing the environmental footprint associated with mining operations. 

Reusing waste materials to achieve desirable properties is essential for a circular economy. Computational modeling plays a key role in optimizing mix designs for sustainable construction materials, expediting the discovery of efficient material combinations \cite{Ginga2020}. For example, Liu et al. \cite{Liu2022} applied machine learning (ML) techniques, including Support Vector Regression (SVR) and Gradient Boosting, to enhance asphalt mixture design. Trained on Superpave mix-design data, these models predicted critical parameters like binder content with high accuracy, demonstrating the practical value of ML in optimizing mix designs for properties such as compressive strength or porosity.

The lack of comprehensive datasets is a prevalent issue in the stone/marble industry. This scarcity necessitates the completion of datasets through predictive techniques such as linear regression or averaging mean values, thereby achieving a sufficiently robust dataset for subsequent analysis. The initial deficit in data significantly undermines the efficacy of predictive models, impeding their utility in practical applications. Consequently, this limitation constrains the applicability and reliability of these models in real-world scenarios \cite{cao2023prediction}. A promising approach to mitigate this challenge is the implementation of machine learning models through a meta-learning approach. These models are trained on related datasets, enabling them to generalize across diverse tasks, even in data-constrained environments effectively. Meta-learning leverages prior knowledge to enhance adaptability and performance in new tasks, thereby addressing the limitations imposed by incomplete datasets. By adopting meta-learning techniques, the stone industry can transcend the limitations of data scarcity, transforming incomplete datasets into actionable insights. This approach augments the precision and reliability of predictive models and expands their applicability in real-world contexts, fostering innovation and excellence within the field.



\section{Method}\label{sec:method}

The simulation of a material's specific attribute is a critical task in the material sciences, geology, and engineering, as it directly influences the physical and chemical properties of the material. However, traditional methods of determining such properties often rely on empirical models and experimental data, which can be time-consuming and costly. For example, conventional mercury intrusion porosimetry techniques or gas absorption require extensive laboratory setups, careful sample preparation, and lengthy testing procedures. These methods also consume significant amounts of materials and reagents, further driving up costs. In this section, we introduce a simulator guided by the machine learning to simulate the porosity of the material. The training and testing process of the simulator is divided into four stages, as reported in Figure 2.

\begin{figure}[h!]
    \centering
    \includegraphics[width=0.8\linewidth]{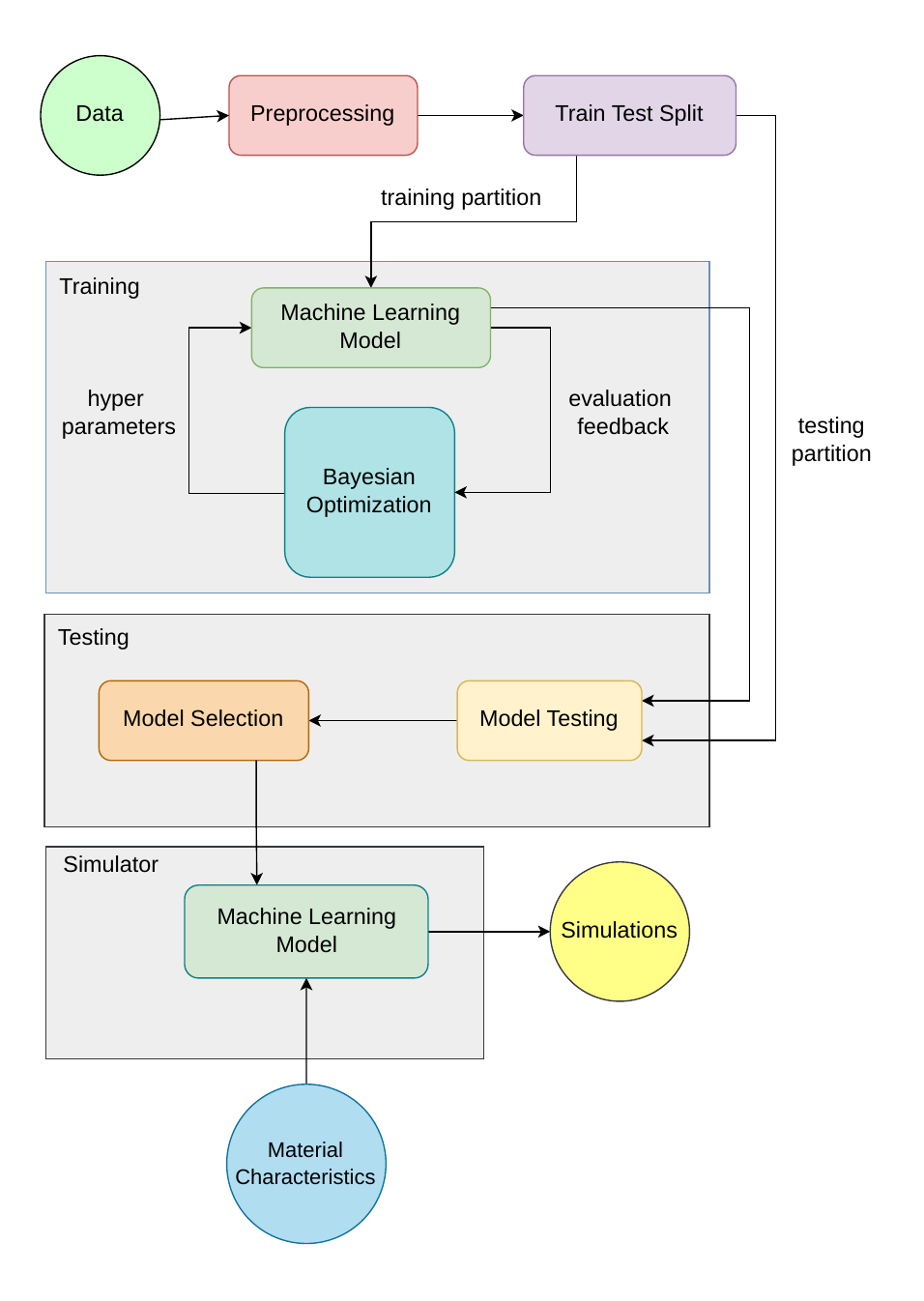}
    \caption{The end-to-end workflow of a simulator based on the machine learning model development and deployment process. It is divided into four main phases: Preprocessing, Training, Testing, and Simulation, with the ultimate goal of producing simulations based on material characteristics. The circle represents the data, while the rectangle represents the process.}
    \label{fig:workflow}
\end{figure}

\subsection{Data Preprocessing}
During the data preprocessing phase, we applied Z-Normalization to standardize the dataset. This method adjusts the features so that they have a mean of zero and a standard deviation of one. Given that certain features in the dataset span a much wider range compared to others, standardization ensures that all features contribute equally to the training process, thereby preventing biased learning. Following this step, we utilized a train-test split technique to divide the data into training and testing sets. The training set was used for model training, while the test set, unseen during the training phase, was reserved for evaluating the model's performance.
\subsection{Training}
In this stage, we train different machine learning algorithms \ie \textit{Gradient Boosting, XGBoost, Random Forest (RF), and Support Vector Machines (SVM)}. Each of these algorithms has unique strengths: Gradient Boosting provides a powerful method for improving predictive accuracy through iterative training; XGBoost and Random Forest offer powerful ensemble methods capable of capturing intricate patterns; and SVM excels in high-dimensional spaces, making it suitable for complex datasets \cite{callens2020using,abdullah2021machine}.

However, determining the optimal hyper-parameters for these machine learning models is challenging for several reasons. First, the hyper-parameter space can be vast and complex, with many possible combinations of values \cite{shahriari2015taking}. For instance, in XGBoost, parameters like learning rate, max depth, and the number of estimators need to be finely tuned, each with a wide range of potential values. Second, hyper-parameter tuning is computationally intensive, often requiring extensive cross-validation to assess the performance of each combination \cite{yu2020hyper}. Third, the interaction between different hyper-parameters can be non-linear and unpredictable, making it challenging to select the best settings intuitively~\cite{yang2020hyperparameter}.

To address these challenges, we employ a specific meta-learning approach that is Bayesian Optimization (BO). Bayesian optimization is a technique that efficiently searches for the best hyper-parameters by building a probabilistic model (a Gaussian Process) of the objective function and using this model to select the most promising hyper-parameters to evaluate next~\cite{du2022bayesian}. This approach automates the tuning process and systematically balances exploration and exploitation, ensuring that the models are finely tuned for the best predictive performance. Doing so improves model accuracy and generalization while adapting the learning algorithms to the specific characteristics of the data.

\subsection{Testing}
In the testing phase, we evaluate the models trained in the previous phase and select the best model for our simulation. So, we tested our models with the test set from the data preprocessing phase quantitatively with the help of numerous evaluation metrics (explained in Section \ref{sec:exps}). After evaluating the models, we select the best-performing model that can be used as a simulator.
\subsection{Simulator}
In this section, we employed a machine-learning-based simulator specifically designed to estimate material porosity using a trained, best-performing model from the previous section. The material characteristics will be the input of the simulator, and based on them, the simulator simulates the material's porosity. Unlike a traditional predictor, which forecasts outcomes based solely on historical data, our simulator offers more comprehensive capabilities. It models and simulates the complex relationships between various material parameters, enabling the estimation of porosity without relying on historical data and using only current material inputs. While predictors focus on making future predictions based on learned patterns from past data, our simulator replicates the underlying interactions within a system, providing a way to experiment with different parameter configurations without the need for physical tests. 

This approach significantly reduces the time and cost associated with extensive experimental procedures. So, as an example and position of this paper, our simulator estimates porosity based on specific material properties (outlined in Table \ref{tab:features}), showcasing its effectiveness as a robust and flexible tool for material analysis and research. This innovation supports rapid prototyping and iterative testing, driving efficiency and advancement in material science.

\section{Experiments}\label{sec:exps}

To validate our idea of using ML to predict a good-quality mixture from stone-cutting sludge, we need data on how these materials are treated to transform from merely stone residual materials to industry-grade ones. Unfortunately, as explained in Section~\ref{sec:related}, the lack of data in these regards significantly reduces the applicability of ML models. However, to demonstrate the potential of the ML approaches in modeling the physical properties and evaluating the effective quality of the resulting mixture, we focus our experiments on data from concrete processing. 
Concrete quality estimation is a very similar task to the one we aim to solve since it's the process of determining the physical properties of a final mixture, which gives some processing applied.

One of the key measures to assess concrete quality is evaluating its \textit{porosity}. Concretes with low porosity are highly desirable for industrial applications due to their enhanced strength, durability, and resistance to environmental degradation. The dataset we utilized focuses on the characteristics of concrete and how each feature impacts the material's quality.  The dataset is publically available \mbox{\cite{cao2023prediction} while the description of the features is presented }in Table \mbox{\ref{tab:features}}

\begin{table}[t]
\renewcommand{\arraystretch}{1.5} 
\centering
\caption{The features were chosen based on their Pearson correlation with porosity, as each property is expected to influence porosity variations. These features were subsequently utilized in the modeling and prediction processes.}
\label{tab:features}
\resizebox{\textwidth}{!}{
\begin{tabular}{|l|p{10cm}|}
\hline
\textbf{Features}& \textbf{Description} \\
\hline
Water to binder Ratio (WB) & Determines the balance of water to cementitious materials.\\
\hline
Binder & Binds aggregates together through hydration.\\
\hline
Flyash & Enhances the concrete's strength and durability while reducing its environmental footprint.\\
\hline
Slag & Improves concrete's durability and resistance to chemical attack.\\
\hline
Superplasticizer & Additives that enhance workability and reduce water content without affecting the WB ratio.\\
\hline
Aggregate & Provides bulk and dimensional stability to concrete mixes.\\
\hline
Curing days & Ensures optimal hydration and concrete strength development over time.\\
\hline
Strength & Encompasses compressive, tensile, and flexural properties of a concrete.\\
\hline
\end{tabular}
}
\end{table}


We randomly split the available data and used 80\% for training our models and the remaining 20\% for validation. 
Several evaluation metrics, such as Mean Squared Error (MSE), the R$^2$, and the Mean Absolute Percentage Error (MAPE) are used to evaluate the machine learning models quantitatively. 
As models, we used five machine learning models: support vector regression (SVR), Linear Regressor (LR), random forest (RF), gradient boosting (GB), and XGboost (XGB). The results are shown in Table~\ref{tab:results}.

\begin{table}[t]
\renewcommand{\arraystretch}{1.5} 
\centering
\caption{Results of the different models on the dataset for predicting concrete's porosity. The results in terms of MSE, R$^2$, and MAPE are presented. The $\downarrow$ symbol indicates that lower values are better, while the $\uparrow$ symbol indicates that higher values are better.}
\label{tab:results}
\resizebox{\textwidth}{!}{
\begin{tabular}{|l|c|c|c|c|c|c|}
\hline
\textbf{Model} & \textbf{Train MSE} $\downarrow$ & \textbf{Test MSE} $\downarrow$ & \textbf{Train R$^2$} $\uparrow$ & \textbf{Test R$^2$} $\uparrow$ & \textbf{Train MAPE} $\downarrow$ & \textbf{Test MAPE} $\downarrow$ \\
\hline
GB & 0.5287 & 0.6946 & 0.4738 & 0.2820 & 1.0612 & 0.9051 \\
\hline
SVM & 0.2367 & 0.3734 & 0.7644 & 0.6141 & 0.7867 & 0.8529 \\
\hline
LR & 0.3108 & 0.3119 & 0.6907 & 0.6776 & 2.0035 & 1.8318 \\
\hline
RF & 0.0262 & 0.1683 & 0.9739 & 0.8260 & 0.4367 & 0.7098 \\
\hline
XGB & \textbf{0.0164} & \textbf{0.1623} & \textbf{0.9837} & \textbf{0.8323} & \textbf{0.3824} & \textbf{0.5590} \\
\hline
\end{tabular}
}
\end{table}
To further enhance our models' predictive capabilities and robustness across diverse tasks and environments, we integrated a meta-learning optimization technique known as Bayesian Optimization. This approach efficiently searches for and adapts optimal hyper-parameters from the complex and large hyper-parameter space of machine learning models. Specifically, it optimizes both hyper-parameters, which control the learning process, and model architecture parameters, which define the structure of the machine learning model. A well-defined search space is also needed for the optimization, hence we selected the spaces shown in Table \ref{tab:hyper-parameters}.

\begin{table}[t]
\renewcommand{\arraystretch}{1.5} 
\centering
\caption{Hyper-parameters search spaces used in the meta-learning.}
\resizebox{\textwidth}{!}{  
\label{tab:hyper-parameters}
\begin{tabular}{|c|c|c|}
\hline
\textbf{Models} & \textbf{Hyper-parameters}& \textbf{Search Space} \\
\hline
\multirow{4}{*}{\textbf{SVM}} & C & (1e-6, 1e+6) \\
    & Epsilon & (1e-6, 1e+1) \\
    & Kernel & (['linear', 'poly', 'rbf', 'sigmoid']) \\
    & Degree & (1, 5) \\
\hline
\multirow{3}{*}{\textbf{Gradient Boosting}} & N\_estimators & (10, 150) \\
    & Learning\_rate & (1e-6, 1e-1) \\
    & Max\_depth & (1, 250) \\
    & Min\_sample\_leaf & (1,50)\\
    & Min\_samples\_split & (2, 50) \\
\hline
\multirow{4}{*}{\textbf{Random Forest}} & N\_estimators & (10, 250) \\
    & Min\_samples\_leaf & (1, 50) \\
    & Min\_samples\_split & (2, 50) \\
    & Max\_depth & (10, 200) \\
\hline
\multirow{3}{*}{\textbf{XGB Regressor}} & Max\_depth & (1, 180) \\
    & Learning\_rate & (1e-7, 1e-1) \\
    & Alpha & (1e-7, 1e+1) \\
    & N\_estimators & (10, 250) \\
\hline
\end{tabular}
}
\end{table}
Bayesian optimization played a crucial role in generalizing from limited data, enabling the development of adaptive algorithms that quickly assimilate new information and adjust behavior accordingly. This method improved the efficiency of all four models except Linear Regression, which doesn't have hyper-parameters to optimize while also providing deeper insights into the underlying principles of learning dynamics.

\begin{table}[t]
\renewcommand{\arraystretch}{1.5} 
\centering
\caption{Results of the different models on the dataset using meta-learning. The results in terms of MSE, R$^2$, and MAPE are presented. The $\downarrow$ symbol indicates that lower values are better, while the $\uparrow$ symbol indicates that higher values are better.}
\label{tab:meta-learning}
\resizebox{\textwidth}{!}{
\begin{tabular}{|l|c|c|c|c|c|c|}
\hline
\textbf{Model}  & \textbf{Train MSE} $\downarrow$ & \textbf{Test MSE} $\downarrow$ &\textbf{Train R$^2$} $\uparrow$ & \textbf{Test R$^2$} $\uparrow$ & \textbf{Train MAPE} $\downarrow$ & \textbf{Test MAPE} $\downarrow$\\
\hline
RF & 0.0158 & 0.1677 & 0.9842 & 0.8266 & 0.3642 & 0.5694 \\
\hline
SVM & 0.0584 & 0.1373 & 0.9419 & 0.8581 & 0.8814 & 0.8443 \\
\hline
XGB & 0.0147 & 0.1258 & 0.9854 & 0.8700 & 0.4634 & 0.7749 \\
\hline
GB & \textbf{0.0147} & \textbf{0.1077} & \textbf{0.9854} & \textbf{0.8887} & \textbf{0.4480} & \textbf{0.6933} \\
\hline
\end{tabular}
}
\end{table}

\subsection{Discussion}
Figure~\ref{fig:Fig4} displays the performance of the two best and the worst models by plotting data points across the distribution before any hyper-parameter optimization. The XGB Regressor and Random Forest (RF) models are the top performers, achieving (R$^2$=0.8322 and 0.8260)  values on the testing data, respectively. This indicates a strong correlation between the model predictions and the actual porosity values, explaining approximately 83\% and 82\% of the variance in the data. Both models show an excellent fit on the training set and strong performance on the testing set, with the XGB Regressor performing slightly better due to lower error metrics and higher R$^2$ values (see Section~\ref{tab:results}), indicating robust performance with minimal overfitting. RF also demonstrates high performance, as evidenced by the tight clustering around the ideal prediction line, suggesting good accuracy. However, it shows a slightly higher error on the testing set, indicating marginally less generalization than the XGB Regressor. In contrast, GB performs the worst among the models, with an (R$^2$ = 0.4738), showing significant deviations in predictions and poor generalization to the testing set. The SVM model performs moderately well, indicating decent predictive capability but with higher error and a lower (R$^2$=0.614) than RF and XGB Regressor. LR underperforms compared to the more complex models, indicating it may struggle to capture non-linear patterns in the data.
\begin{figure}[h!]
    \centering
    \includegraphics[width=\linewidth]{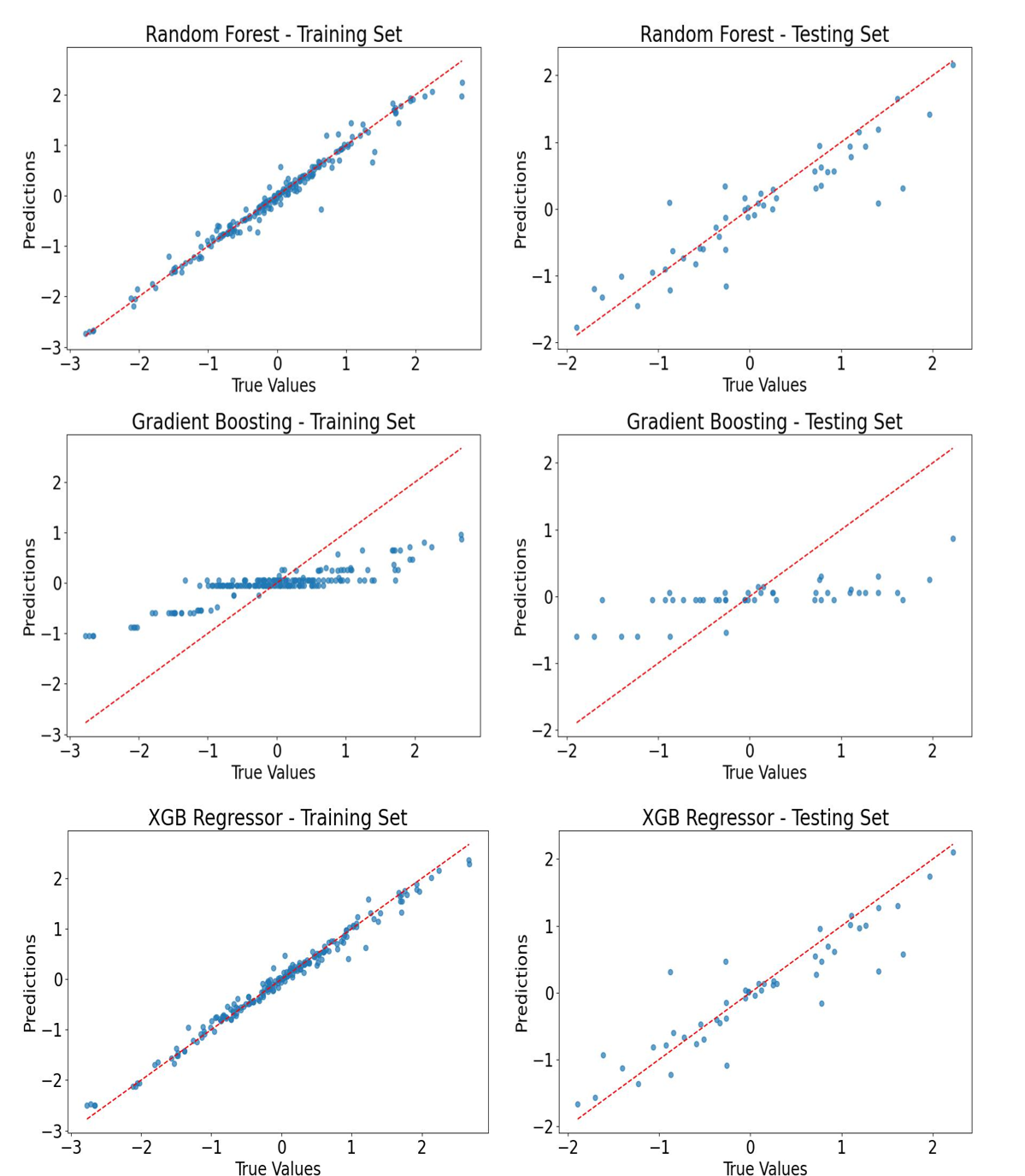}
    \caption{The plot shows the performance of different models on training and testing sets, respectively. The x-axis represents the true values, while the y-axis represents the predictions made by the model without meta-learning. The blue points are individual data points, and the red dashed line represents the ideal line where the predictions perfectly match the true values.}
      \label{fig:Fig3}
\end{figure}

After tuning the models using Bayesian optimization, all models showed significant improvements in their performance metrics, as depicted in Section~\ref{tab:meta-learning}, and the two best models are displayed in Figure~\ref{fig:Fig3}. GB exhibited the most notable enhancements, reducing its test MSE and increasing its test R$^2$ substantially from 0.4738 to 0.8887. This improvement is likely due to GB's initial tendency to capture local trends well but struggle with global trends without proper hyper-parameter tuning. Bayesian optimization significantly enhanced GB's performance, achieving a much better fit and predictive accuracy on the test data. Post-optimization, GB improved its ability to explain the variance in the data and reduce prediction errors, making it the best-performing model among those optimized. The XGBoost model also saw significant improvements, with its R$^2$ increasing from 0.832 to 0.887 on the test dataset, enhancing its effectiveness and generalizability. The SVM model significantly reduced its test MSE and increased its test R$^2$, indicating better performance and predictive accuracy. Meanwhile, the RF model maintained strong performance with slight improvements in both its training and test metrics, showcasing its reliability and efficiency in handling the data. Overall, while all models improved, GB emerged as the best performer, followed closely by XGBoost.
\begin{figure}[!h]
    \centering
    \includegraphics[width=\linewidth]{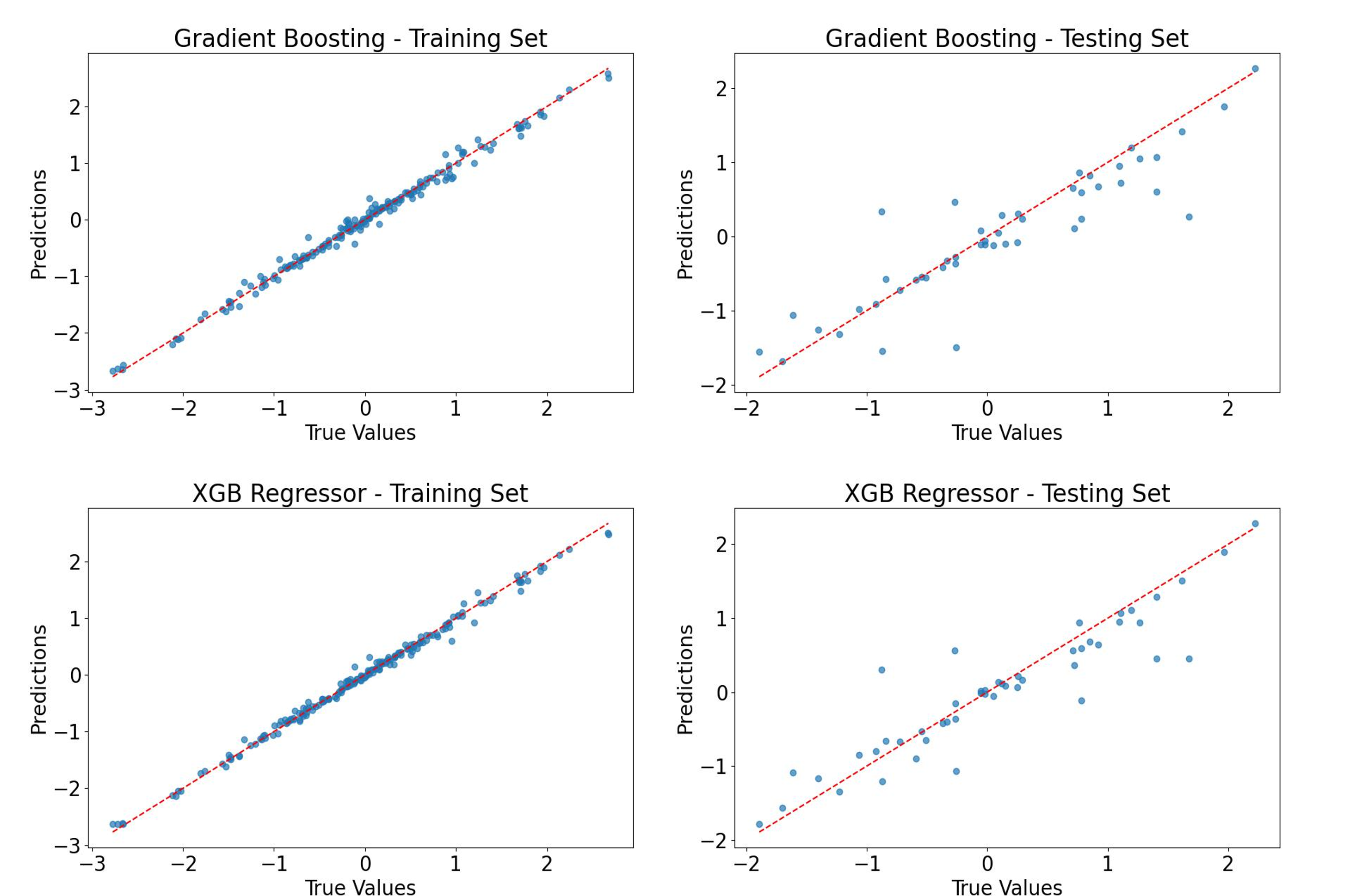}
    \caption{The plot displays the best model's performance, with meta-learning through Bayesian Optimization.}
    \label{fig:Fig4}
\end{figure}

These results demonstrate that applying meta-learning through Bayesian optimization can effectively tune models for predicting porosity with high accuracy in this scenario. Although this approach is not limited to predicting porosity and can be generalized to determine various material properties, certain limitations must be addressed before applying this model. Firstly, the model's accuracy may significantly decrease if the data is overly noisy or if there are insufficient features relative to the complexity of the property being predicted. Noisy data consists of random or unpredictable fluctuations that obscure underlying patterns or relationships, thus complicating the identification of target features. These fluctuations can distort the relationships among variables, reducing the model's reliability. Additionally, an insufficient number of features can negatively impact prediction accuracy. When predicting properties with physical or chemical significance, multiple interacting factors typically contribute to the outcome. A comprehensive feature set is essential to capture these factors accurately and to improve predictive performance. Therefore, it is crucial to ensure that all relevant features are available and accounted for when developing models for these types of predictions.

\section{Conclusions}\label{sec:conclusions}
This position paper highlights the pressing need for innovative solutions to address the environmental and economic challenges posed by stone-cutting sludge, particularly in the context of the natural stone industry. The substantial volume of waste generated annually, coupled with the high costs of disposal, underscores the urgency of developing sustainable alternatives to landfilling.

By proposing the integration of artificial intelligence and machine learning (ML) techniques in the mix-design process, we present a promising pathway toward large-scale upcycling of marble dust in construction materials. The application of ML techniques, as demonstrated with the concrete dataset, showcases the potential for significantly improving prediction accuracy in material properties like porosity. 
This leap in predictive capability could revolutionize the way we approach mix-design, moving from resource-intensive empirical methods to efficient, simulation-driven processes.
The data on the stone-cutting sludge are not available easily, as explained in Section~\ref{sec:related}, but in this position paper, we propose that this strategy could be effectively applied to achieve improved results with smaller datasets of this type.

The advantages of the proposed approach are multifaceted. Firstly, it offers a cost-effective solution by dramatically reducing the need for extensive physical experiments. Secondly, it accelerates the development process, allowing for rapid iteration and optimization of material compositions. Thirdly, it provides a means to easily adapt mix-designs to varying regulatory standards across different regions, enhancing compliance and applicability.


In essence, this position paper advocates for a paradigm shift in how we approach waste management and material design in the stone industry. By harnessing the power of artificial intelligence, we can pave the way for more sustainable, efficient, and economically viable practices, ultimately contributing to a greener and more circular future in construction and beyond.

%
%
%
\bibliographystyle{splncs04}
\bibliography{main.bib}
\end{document}